\documentclass[11pt,a4paper]{article}
\usepackage[hyperref]{acl2017}
\usepackage{times}
\usepackage{latexsym}
\usepackage{url}
\usepackage{graphicx}
\usepackage{subfigure}

\usepackage{algorithm}
\usepackage{algorithmic}
\usepackage{booktabs}
\usepackage{multirow}
\usepackage{array}
\usepackage{color}
\usepackage{amsmath,amsfonts,amssymb,bbm,epsfig,bm}
\usepackage{float}
\usepackage{multirow}
\usepackage{balance}

\usepackage[titletoc,title]{appendix}

\aclfinalcopy 
\def\aclpaperid{147} 

\newcolumntype{L}[1]{>{\raggedright\arraybackslash}p{#1}}
\newcolumntype{R}[1]{>{\raggedleft\arraybackslash}p{#1}}
\newcolumntype{?}{!{\vrule width 1pt}}

\makeatletter
\newcommand{\thickhline}{%
    \noalign {\ifnum 0=`}\fi \hrule height 1pt
    \futurelet \reserved@a \@xhline
}
\newcolumntype{"}{@{\hskip\tabcolsep\vrule width 1pt\hskip\tabcolsep}}
\makeatother

\newcommand{\E}{\mathbb{E}}

\usepackage{color}

\title{Adversarial Connective-exploiting Networks for \\ Implicit Discourse Relation Classification}

\author{Lianhui Qin$^1$,~ Zhisong Zhang$^1$,~ Hai Zhao$^1$,~ Zhiting Hu$^2$,~ Eric P. Xing$^2$\\
$^1$Shanghai Jiao Tong University,\quad $^2$Carnegie Mellon University \\
  {\tt \{qinlianhui, zzs2011\}@sjtu.edu.cn,~ zhaohai@cs.sjtu.edu.cn,}\\ 
  {\tt \{zhitingh, epxing\}@cs.cmu.edu} \\ 
  \\}

\date{}

\begin{document}

\maketitle
\begin{abstract}
Implicit discourse relation classification is of great challenge due to the lack of connectives as strong linguistic cues, which motivates the use of annotated implicit connectives to improve the recognition.
We propose a feature imitation framework in which an implicit relation network is driven to learn from another neural network with access to connectives, and thus encouraged to extract similarly salient features for accurate classification. We develop an adversarial model to enable an adaptive imitation scheme through competition between the implicit network and a rival feature discriminator. Our method effectively transfers discriminability of connectives to the implicit features, and achieves state-of-the-art performance on the PDTB benchmark. 
\end{abstract}

\section{Introduction}

Discourse relations connect linguistic units such as clauses and sentences to form coherent semantics. Identification of discourse relations
can benefit a variety of downstream applications including question answering \cite{D13-1070}, machine translation \cite{P14-2047}, text summarization \cite{D14-1168}, and so forth.

Connectives (e.g., {\it but}, {\it so}, etc) are one of the most critical linguistic cues for identifying discourse relations. When explicit connectives are present in the text, a simple frequency-based mapping is sufficient to achieve over 85\% classification accuracy~\cite{xue-EtAl:2016:CoNLL-ST}. 
In contrast, {\it implicit discourse relation recognition} has long been seen as a challenging problem, with the best accuracy so far still lower than 50\%. In the implicit case, discourse relations are not lexicalized by connectives, but to be inferred from relevant sentences (i.e., {\it arguments}). For example, the following two adjacent sentences Arg1 and Arg2 imply relation {\it Cause} (i.e., Arg2 is the cause of Arg1). 
\begin{itemize}
\setlength\itemsep{0.2pt}
\item[] [{\it Arg1}]: Never mind. 
\item[] [{\it Arg2}]: You already know the answer.
\item[] [{\it Implicit connective}]: Because
\item[] [{\it Discourse relation}]: Cause
\end{itemize}
Various attempts have been made to directly infer underlying relations by modeling the semantics of the arguments, ranging from feature-based methods~\cite{lin2009recognizing,pitler-louis-nenkova:2009:ACLIJCNLP} to the very recent end-to-end neural models~\cite{chen2016implicit,qin2016stacking}. 
Despite impressive performance, the absence of strong explicit connective cues has made the inference extremely hard and hindered further improvement. In fact, even the human annotators would make use of connectives to aid relation annotation. For instance, the popular Penn Discourse Treebank (PDTB) benchmark data~\cite{prasad2008penn} was annotated by first inserting a connective expression (i.e., {\it implicit connective}, as shown in the above example) manually, and determining the abstract relation by combining both the implicit connective and contextual semantics.

Therefore, the huge performance gap between explicit and implicit parsing (namely, 85\% vs 50\%), as well as the human annotation practice, strongly motivates to incorporate connective information to guide the reasoning process. 
This paper aims to advance implicit parsing by making use of annotated implicit connectives available in training data.
Few recent work has explored such combination. 
\newcite{zhou2010predicting} developed a two-step approach by first predicting implicit connectives whose sense is then disambiguated to obtain the relation. However, the pipeline approach usually suffers from error propagation, and the method itself has relied on hand-crafted features which do not necessarily generalize well. Other research leveraged explicit connective examples for data augmentation~\cite{rutherford2015improving,braud2015comparing,ji2015closing,braud2016learning}. Our work is orthogonal and complementary to this line.

In this paper, we propose a novel neural method that incorporates implicit connectives in a principled {\it adversarial} framework. We use deep neural models for relation classification, and take the intuition that, sentence arguments integrated with connectives would enable highly discriminative neural features for accurate relation inference, and an ideal implicit relation classifier, even though without access to connectives, should mimic the connective-augmented reasoning behavior by extracting similarly salient features. We therefore setup a secondary network in addition to the implicit relation classifier, building upon connective-augmented inputs and serving as a feature learning model for the implicit classifier to emulate. 

Methodologically, however, feature imitation in our problem is challenging due to the semantic gap induced by adding the connective cues. It is necessary to develop an adaptive scheme to flexibly drive learning and transfer discriminability. We devise a novel adversarial approach which enables a self-calibrated imitation mechanism. 
Specifically, we build a discriminator which distinguishes between the features by the two counterpart networks. The implicit relation network is then trained to correctly classify relations and simultaneously to fool the discriminator, resulting in an adversarial framework.
The adversarial mechanism has been an emerging method in different context, especially for image generation~\cite{goodfellow2014generative} and domain adaptation~\cite{ganin2016domain,chen2016adversarial}. Our adversarial framework is unique to address neural feature emulation between two models. Besides, to the best of our knowledge, this is the first adversarial approach in the context of discourse parsing. Compared to previous connective exploiting work~\cite{zhou2010predicting,xu2012connective}, our method provides a new integration paradigm and an end-to-end procedure that avoids inefficient feature engineering and error propagation.

We evaluate our method on the PDTB 2.0 benchmark in a variety of experimental settings. The proposed adversarial model greatly improves over standalone neural networks and previous best-performing approaches. We also demonstrate that our implicit recognition network successfully imitates and extracts crucial hidden representations.

We begin by briefly reviewing related work in section~\ref{sec:related}. Section~\ref{sec:model} presents the proposed adversarial model. Section~\ref{sec:exp} shows substantially improved experimental results over previous methods. Section~\ref{sec:conclude} discusses extensions and future work.

\section{Related Work}\label{sec:related}

\subsection{Implicit Discourse Relation Recognition}
There has been a surge of interest in implicit discourse parsing since the release of PDTB~\cite{prasad2008penn}, the first large discourse corpus distinguishing implicit examples from explicit ones. A large set of work has focused on direct classification based on observed sentences, including structured methods with linguistically-informed features~\cite{lin2009recognizing,pitler-louis-nenkova:2009:ACLIJCNLP,zhou2010predicting}, end-to-end neural models~\cite{qin2016shallow,qin2016stacking,chen2016implicit,D16-1130}, and combined approaches~\cite{ji2015one,ji2016latent}. However, the lacking of connective cues makes learning purely from contextual semantics full of challenges.

Prior work has attempted to leverage connective information. 
\newcite{zhou2010predicting} also incorporate implicit connectives, but in a pipeline manner by first predicting the implicit connective with a language model and determining discourse relation accordingly. Instead of treating implicit connectives as intermediate prediction targets which can suffer from error propagation, we use the connectives to induce highly discriminative features to guide the learning of an implicit network, serving as an adaptive regularization mechanism for enhanced robustness and generalization. Our framework is also end-to-end, avoiding costly feature engineering.
Another notable line aims at adapting explicit examples for data synthesis~\cite{biran2013aggregated,rutherford2015improving,braud2015comparing,ji2015closing}, multi-task learning~\cite{lan2013leveraging,liu2016implicit}, and word representation~\cite{braud2016learning}. Our work is orthogonal and complementary to these methods, as we use implicit connectives which have been annotated for implicit examples.

\begin{figure*}[!h]
	\centering
	\includegraphics[width=\textwidth]{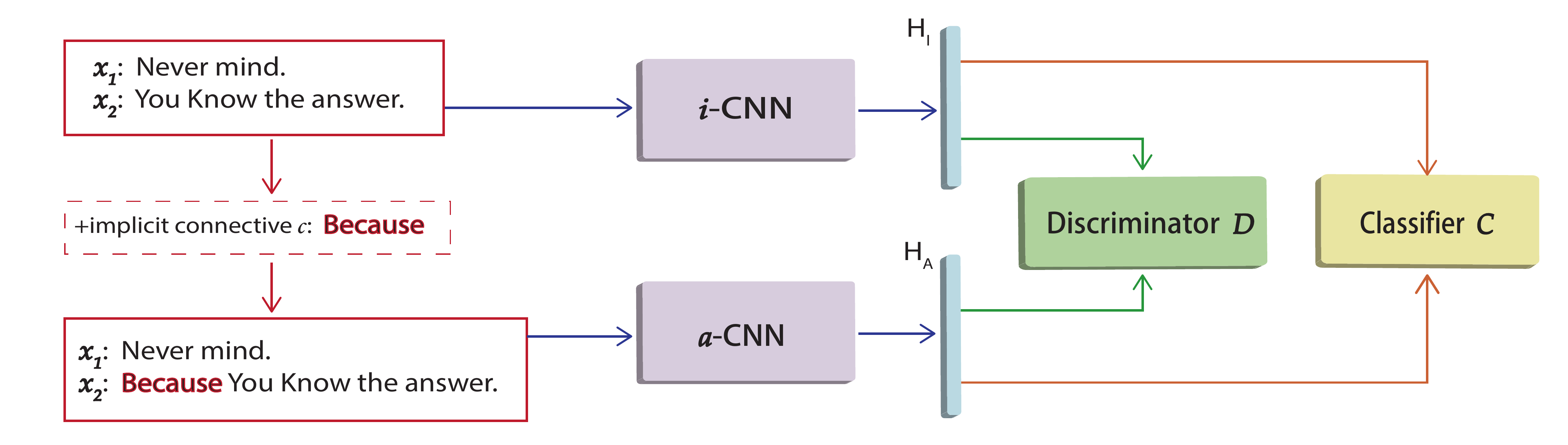}
	\caption{Architecture of the proposed method. The framework contains three main components: 1) an implicit relation network {\it i-CNN} over raw sentence arguments, 2) a connective-augmented relation network {\it a-CNN} whose inputs are augmented with implicit connectives, and 3) a discriminator distinguishing between the features by the two networks. The features are fed to the final classifier for relation classification. The discriminator and {\it i-CNN} form an adversarial pair for feature imitation. At test time, the implicit network {\it i-CNN} with the classifier is used for prediction.}
	\label{fig:model}
\end{figure*}

\subsection{Adversarial Networks}
Adversarial method has gained impressive success in deep generative modeling~\cite{goodfellow2014generative} and domain adaptation~\cite{ganin2016domain}. Generative adversarial nets~\cite{goodfellow2014generative} learn to produce realistic images through competition between an image generator and a real/fake discriminator. Professor forcing~\cite{lamb2016professor} applies a similar idea to improve long-term generation of a recurrent neural language model. Other approaches~\cite{chen2016infogan,hu2017controllable} extend the framework for controllable image/text generation. \newcite{li2015generative,salimans2016improved} propose feature matching which trains generators to match the statistics of real/fake examples. Their features are extracted by the discriminator rather than the classifier networks as in our case. Our work differs from the above since we consider the context of discriminative modeling. Adversarial domain adaptation forces a neural network to learn domain-invariant features using a classifier that distinguishes the domain of the network's input data based on the hidden feature. Our adversarial framework is distinct in that besides the implicit relation network we construct a second neural network serving as a teacher model for feature emulation. 

To the best of our knowledge, this is the first to employ the idea of adversarial learning in the context of discourse parsing. We propose a novel connective exploiting scheme based on feature imitation, and to this end derive a new adversarial framework, achieving substantial performance gain over existing methods. The proposed approach is generally applicable to other tasks for utilizing any indicative side information. We give more discussions in section~\ref{sec:conclude}.

\section{Adversarial Method}\label{sec:model}
Discourse connectives are key indicators for discourse relation. In the annotation procedure of the PDTB implicit relation benchmark, annotators inserted implicit connective expressions  between adjacent sentences to lexicalize abstract relations and help with final decisions. Our model aims at making full use of the provided implicit connectives at training time to regulate learning of implicit relation recognizer, encouraging extraction of highly discriminative semantics from raw arguments, and improving generalization at test time. Our method provides a novel adversarial framework that leverages connective information in a flexible adaptive manner, and is efficiently trained end-to-end through standard back-propagation.

The basic idea of the proposed approach is simple. We want our implicit relation recognizer, which predicts the underlying relation of sentence arguments without discourse connective, to have prediction behaviors close to a {\it connective-augmented} relation recognizer which is provided with a discourse connective in addition to the arguments. The connective-augmented recognizer is in analogy to an annotator with the help of connectives as in the human annotation process, and the implicit recognizer would be improved by learning from such an ``informed'' annotator. 
Specifically, we want the latent features extracted by the two models to match as closely as possible, which explicitly transfers the discriminability of the connective-augmented representations to implicit ones. 

To this end, instead of manually selecting a closeness metric, we take advantage of the adversarial framework by constructing a two-player zero-sum game between the implicit recognizer and a rival discriminator. The discriminator attempts to distinguish between the features extracted by the two relation models, while the implicit relation model is trained to maximize the accuracy on implicit data, and at the same time to confuse the discriminator. 

In the next we first present the overall architecture of the proposed approach (section~\ref{sec:arch}), then develop the training procedure (section~\ref{sec:training}). The components are realized as deep (convolutional) neural networks, with detailed modeling choices discussed in section~\ref{sec:components}.

\subsection{Model Architecture} \label{sec:arch}
Let $(\bm{x}, y)$ be a pair of input and output of implicit relation classification, where $\bm{x}=(\bm{x}_1, \bm{x}_2)$ is a pair of sentence arguments, and $y$ is the underlying discourse relation. Each training example also includes an annotated implicit connective $c$ that best expresses the relation. Figure~\ref{fig:model} shows the architecture of our framework.

The neural model for implicit relation classification ({\it i-CNN} in the figure) extracts latent representation from the arguments, denoted as $H_{I}(\bm{x}_1, \bm{x}_2)$, and feeds the feature into a classifier $C$ for final prediction $C(H_{I}(\bm{x}_1, \bm{x}_2))$. For ease of notation, we will also use $H_{I}(\bm{x})$ to denote the latent feature on data $\bm{x}$.

The second relation network ({\it a-CNN}) takes as inputs the sentence arguments along with an implicit connective, to induce the connective-augmented representation $H_{A}(\bm{x}_1, \bm{x}_2, c)$, and obtains relation prediction $C(H_{A}(\bm{x}_1, \bm{x}_2, c))$. Note that the same final classifier $C$ is used for both networks, so that the feature representations by the two networks are ensured to be within the same semantic space, enabling feature emulation as presented shortly.

We further pair the implicit network with a rival discriminator $D$ to form our  adversarial game. The discriminator is to differentiate between the reasoning behaviors of the implicit network {\it i-CNN} and the augmented network {\it a-CNN}. Specifically, $D$ is a binary classifier that takes as inputs a latent feature $H$ derived from either {\it i-CNN} or {\it a-CNN} given appropriate data (where implicit connectives is either missing or present, respectively). The output $D(H)$ estimates the probability that $H$ comes from the connective-augmented {\it a-CNN} rather than {\it i-CNN}.

\subsection{Training Procedure}\label{sec:training}
The system is trained through an alternating optimization procedure that updates the components in an interleaved manner. In this section, we first present the training objective for each component, and then give the overall training algorithm. 

Let $\bm{\theta}_D$ denote the parameters of the discriminator. The training objective of $D$ is straightforward, i.e., to maximize the probability of correctly distinguishing the input features:
\begin{equation}\label{eq:d}
\small
\begin{split}
\max_{\bm{\theta}_D} \mathcal{L}_{D} = \E_{(\bm{x},c,y)\sim \text{data}} \Big[ &\log D(H_A(\bm{x},c); \bm{\theta}_D)  + \\
&\log (1 - D(H_I(\bm{x}); \bm{\theta}_D)) \Big],
\end{split}
\end{equation}
where $\E_{(\bm{x},c,y)\sim \text{data}} [\cdot]$ denotes the expectation in terms of the data distribution.

We denote the parameters of the implicit network {\it i-CNN} and the classifier {\it C} as $\bm{\theta}_I$ and $\bm{\theta}_{C}$, respectively. The model is then trained to (a) correctly classify relations in training data and (b) produce salient features close to connective-augmented ones. The first objective can be fulfilled by minimizing the usual cross-entropy loss:
\begin{equation}\label{eq:icnn-cls}
\small
\begin{split}
\mathcal{L}_{I,C}(\bm{\theta}_I, \bm{\theta}_C) = \E_{(\bm{x},y)\sim \text{data}} \Big[ J\big(C(H_{I}(\bm{x}; \bm{\theta}_I); \bm{\theta}_C), y\big) \Big],
\end{split}
\end{equation}
where $J(\bm{p}, y)=-\sum_{k}\mathbb{I}(y=k)\log p_k$ is the cross-entropy loss between predictive distribution $\bm{p}$ and ground-truth label $y$. We achieve objective (b) by minimizing the discriminator's chance of correctly telling apart the features:
\begin{equation}\label{eq:icnn-gan}
\small
\begin{split}
\mathcal{L}_{I}(\bm{\theta}_I) = \E_{\bm{x}\sim \text{data}} \Big[ \log \big(1 - D(H_I(\bm{x}; \bm{\theta}_I)) \big) \Big].
\end{split}
\end{equation}

\begin{algorithm}[t]
\centering
\caption{\small Adversarial Model for Implicit Recognition}
\label{alg:opt}
\begin{algorithmic}[1]
\REQUIRE Training data $\{(\bm{x}, c, y)_n\}$\\
\quad\ \  Parameters: $\lambda_1, \lambda_2$  -- balancing parameters \\
\STATE Initialize $\{\bm{\theta}_I, \bm{\theta}_C\}$ and $\{\bm{\theta}_A\}$ by minimizing Eq.\eqref{eq:icnn-cls} and Eq.\eqref{eq:acnn}, respectively
\REPEAT
    \STATE Train the discriminator through Eq.\eqref{eq:d}
    \STATE Train the relation models through Eq.\eqref{eq:all-nn}
\UNTIL{convergence}
\ENSURE Adversarially enhanced implicit relation network {\it i-CNN} with classifier {\it C} for prediction
\end{algorithmic}
\end{algorithm}

The parameters of the augmented network {\it a-CNN}, denoted as $\bm{\theta}_A$, can be learned by simply fitting to the data, i.e., minimizing the cross-entropy loss as follows:
\begin{equation}\label{eq:acnn}
\small
\begin{split}
\mathcal{L}_{A}(\bm{\theta}_A) = \E_{(\bm{x},c,y)\sim \text{data}} \Big[ J\big(C(H_{A}(\bm{x}, c; \bm{\theta}_A)), y\big) \Big].
\end{split}
\end{equation}
As mentioned above, here we use the same classifier $C$ as for the implicit network, forcing a unified feature space of both networks. We combine the above objectives Eqs.\eqref{eq:icnn-cls}-\eqref{eq:acnn} of the relation classifiers and minimize the joint loss:
\begin{equation}\label{eq:all-nn}
\small
\begin{split}
\min_{\bm{\theta}_I, \bm{\theta}_A, \bm{\theta}_C} \mathcal{L}_{I,A,C} = \mathcal{L}_{I,C}(\bm{\theta}_I, \bm{\theta}_C) + \lambda_1 \mathcal{L}_{I}(\bm{\theta}_I) + \lambda_2 \mathcal{L}_{A}(\bm{\theta}_A),
\end{split}
\end{equation}
where $\lambda_1$ and $\lambda_2$ are two balancing parameters calibrating the weights of the classification losses and the feature-regulating loss. In practice, we pretrain the implicit and augmented networks independently by minimizing Eq.\eqref{eq:icnn-cls} and Eq.\eqref{eq:acnn}, respectively. In the adversarial training process, we found setting $\lambda_2=0$ gives stable convergence. That is, the connective-augmented features are fixed after the pre-training stage.

Algorithm~\ref{alg:opt} summarizes the training procedure, where we interleave the optimization of Eq.\eqref{eq:d} and Eq.\eqref{eq:all-nn} at each iteration. More practical details are provided in section~\ref{sec:exp}.
We instantiate all modules as neural networks (section~\ref{sec:components}) which are differentiable, and perform the optimization efficiently through standard stochastic gradient descent and back-propagation.

Through Eq.\eqref{eq:d} and Eq.\eqref{eq:icnn-gan}, the discriminator and the implicit relation network follow a minimax competition, which drives both to improve until the implicit feature representations are close to the connective-augmented latent representations, encouraging the implicit network to extract highly discriminative features from raw sentence arguments for relation classification. Alternatively, we can see Eq.\eqref{eq:icnn-gan} as an {\it adaptive} regularization on the implicit model, which, compared to pre-fixed regularizors such as $\ell_2$-regularization, provides a more flexible, self-calibrated mechanism to improve generalization ability.

\begin{figure}[t]
	\centering
	\includegraphics[width=0.48\textwidth]{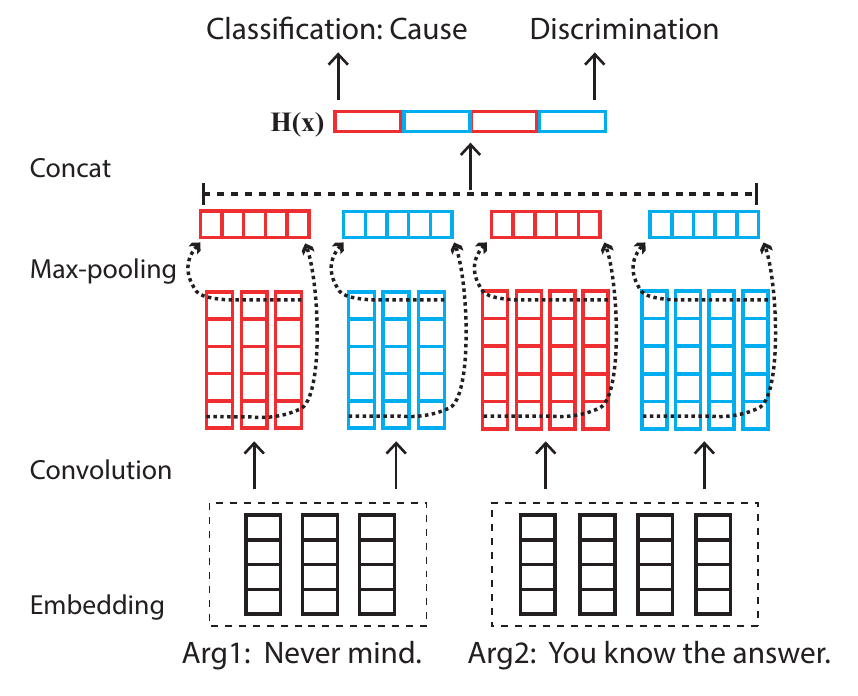}
	\caption{Neural structure of {\it i-CNN}. Two sets of convolutional filters are shown, with the corresponding features in red and blue, respectively. The weights of the filters on two input arguments are tied.}
	\label{fig:icnn}
\end{figure}
\begin{figure}[!h]
	\centering
    \hspace{-20pt}
	\includegraphics[width=0.4\textwidth]{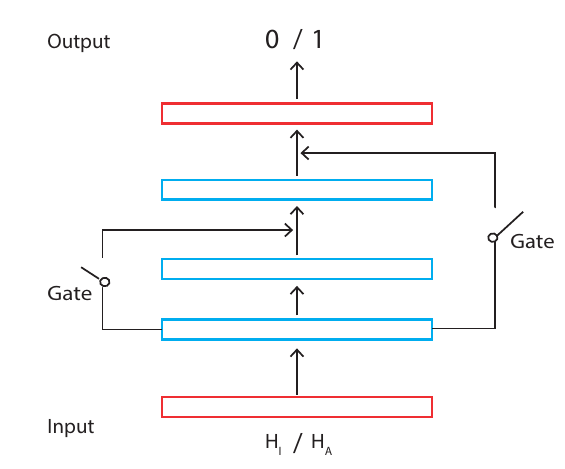}
	\caption{Neural structure of the discriminator {\it D}.}
	\label{fig:d}
\end{figure}

\subsection{Component Structures}\label{sec:components}
We have presented our adversarial framework for implicit relation classification. We now discuss the model realization of each component. All components of the framework are parameterized with neural networks. 
Distinct roles of the modules in the framework lead to different modeling choices. 

\paragraph{Relation Classification Networks}
Figure~\ref{fig:icnn} illustrates the structure of the implicit relation network {\it i-CNN}. We use a convolutional network as it is a common architectural choice for discourse parsing. The network takes as inputs the word vectors of the tokens in each sentence argument, and maps each argument to intermediate features through a shared convolutional layer. The resulting representations are then concatenated and fed into a max pooling layer to select most salient features as the final representation. The final classifier $C$ is a simple fully-connected layer followed by a softmax classifier.

The connective-augmented network {\it a-CNN} has a similar structure as {\it i-CNN}, wherein implicit connective is appended to the second sentence as input. The key difference from {\it i-CNN} is that here we adopt {\it average k-max} pooling, which takes the average of the top-$k$ maximum values in each pooling window.
The reason is to prevent the network from solely selecting the connective induced features (which are typically the most salient features) which would be the case when using max pooling, but instead force it to also attend to contextual features derived from the arguments. This facilitates more homogeneous output features of the two networks, and thus facilitates feature imitation. In all the experiments we fixed $k=2$.

\paragraph{Discriminator}\label{sec:disc}
The discriminator is a binary classifier to identify the correct source of an input feature vector. To make it a strong rival to the feature imitating network ({\it i-CNN}), we model the discriminator as a multi-layer perceptron (MLP) enhanced with gated mechanism for efficient information flow~\cite{srivastava2015highway,qin2016stacking}, as shown in Figure~\ref{fig:d}.

\section{Experiments}\label{sec:exp}
We demonstrate the effectiveness of our approach both quantitatively and qualitatively with extensive experiments. We evaluate prediction performance on the PDTB benchmark in different settings. Our method substantially improves over a diverse set of previous models, especially in the practical multi-class classification task. We perform in-depth analysis of the model behaviors, and show our adversarial framework successfully enables the implicit relation model to imitate and learn discriminative features.

\subsection{Experiment Setup}\label{sec:setup}
We use PDTB 2.0\footnote{http://www.seas.upenn.edu/$\sim$pdtb/}, one of the largest manually annotated discourse relation corpus. The dataset contains 16,224 implicit relation instances in total, with three levels of senses: Level-1 \textit{Class}, Level-2 \textit{Type}, and Level-3 \textit{Subtypes}. The 1st level consists of four major relation \textit{Class}es: \textsc{Comparison}, \textsc{Contingency}, \textsc{Expansion} and \textsc{Temporal}. The 2nd level contains 16 \textit{Type}s.

To make extensive comparison with prior work of implicit discourse relation classification, we evaluate on two popular experimental settings: 1) multi-class classification for 2nd-level types~\cite{lin2009recognizing,ji2015one}, 
and 2) one-versus-others binary classifications for 1st-level classes~\cite{pitler-louis-nenkova:2009:ACLIJCNLP}. We describe the detailed configurations in the following respective sections. We will focus our analysis on the multi-class classification setting, which is most realistic in practice and serves as a building block for a complete discourse parser such as that for the shared tasks of CoNLL-2015 and 2016~\cite{xue2015conll,xue-EtAl:2016:CoNLL-ST}.

\paragraph{Model Training}
We provide detailed model and training configurations in the supplementary materials, and only mention a few of them here. Throughout the experiments {\it i-CNN} and {\it a-CNN} contains 3 sets of convolutional filters with the filter sizes selected on the dev set.
The final single-layer classifier {\it C} contains 512 neurons. The discriminator {\it D} consists of 4 fully-connected layers, with 2 gated pathways from layer 1 to layer 3 and layer 4 (Figure~\ref{fig:d}).

For adversarial model training, it is critical to keep balance between the progress of the two players. We use a simple strategy which at each iteration optimizes the discriminator and the implicit relation network on a randomly-sampled minibatch. We found this is enough to stabilize the training.  The neural parameters are trained using AdaGrad~\cite{duchi2011adaptive} with an initial learning rate of 0.001. For the balancing parameters in Eq.\eqref{eq:all-nn}, we set $\lambda_1=0.1$, while $\lambda_2=0$. That is, after the initialization stage the weights of the connective-augmented network {\it a-CNN} are fixed. This has been shown capable of giving stable and good predictive performance for our system.

\begin{table*}[!h]
  \centering
 \small
\begin{tabular}{l L{6.2cm} L{2cm} L{1.5cm} } 
\cmidrule[\heavyrulewidth]{1-4}
&Model&PDTB-Lin&PDTB-Ji\\
\cmidrule{1-4}
1 & Word-vector &34.07 &36.86\\
2 & CNN & 43.12 &44.51\\
3 & Ensemble & 42.17 & 44.27\\
4 & Multi-task&43.73&44.75\\
5 & $\ell_2$-reg&44.12&45.33\\
\cmidrule{1-4}
6 & \newcite{lin2009recognizing}&40.20&-\\
7 &\newcite{lin2009recognizing}+Brown clusters&-&40.66\\
8 &\newcite{ji2015one}&-&44.59\\
9 &\newcite{C16-1180}&43.81&45.04\\
\cmidrule{1-4}
10 & Ours&\textbf{44.65}&\textbf{46.23}\\
\cmidrule[\heavyrulewidth]{1-4}
\end{tabular}
\caption{Accuracy (\%) on the test sets of the PDTB-Lin and PDTB-Ji settings for \textbf{multi-class classification}. Please see the text for more details.} 
\label{tab:multi}
\end{table*}

\subsection{Implicit Relation Classification}
We will mainly focus on the general multi-class classification problem in two alternative settings adopted in prior work, showing the superiority of our model over previous state of the arts. We perform in-depth comparison with carefully designed baselines, providing empirical insights into the working mechanism of the proposed framework. For broader comparisons we also report the performance in the one-versus-all setting.

\paragraph{Multi-class Classifications}\quad\\
We first adopt the standard PDTB splitting convention following~\cite{lin2009recognizing}, denoted as PDTB-Lin, where sections 2-21, 22, and 23 are used as training, dev, and test sets, respectively. The most frequent 11 types of relations are selected in the task. During training, instances with more than one annotated relation types are considered as multiple instances, each of which has one of the annotations. At test time, a prediction that matches one of the gold types is considered as correct. The test set contains 766 examples. Please refer to \cite{lin2009recognizing} for more details. An alternative, slightly different multi-class setting is used in \cite{ji2015one}, denoted as PDTB-Ji, where sections 2-20, 0-1, and 21-22 are used as training, dev, and test sets, respectively. The resulting test set contains 1039 examples. We also evaluate in this setting for thorough comparisons.

Table~\ref{tab:multi} shows the classification accuracy in both of the settings. We see that our model (Row~10) achieves state-of-the-art performance, greatly outperforming previous methods (Rows~6-9) with various modeling paradigms, including the linguistic feature-based model~\cite{lin2009recognizing}, pure neural methods~\cite{qin2016stacking}, and combined approach~\cite{ji2015one}.

To obtain better insights into the working mechanism of our method, we further compare with a set of carefully selected baselines as shown in Rows 1-5. 1) ``Word-vector'' sums over the word vectors for sentence representation, showing the base effect of word embeddings. 2) ``CNN'' is a standalone convolutional net having the exact same architecture with our implicit relation network. Our model trained within the proposed framework provides significant improvement, showing the benefits of utilizing implicit connectives at training time. 3) ``Ensemble'' has the same neural architecture with the proposed framework except that the input of {\it a-CNN} is not augmented with implicit connectives. This essentially is an ensemble of two implicit recognition networks. We see that the method performs even inferior to the single CNN model. This further confirms the necessity of exploiting connective information.  4) ``Multi-task'' is the convolutional net augmented with an additional task of simultaneously predicting the implicit connectives based on the network features. As a straightforward way of incorporating connectives, we see that the method slightly improves over the stand-alone CNN, while falling behind our approach with a large margin. This indicates that our proposed feature imitation is a more effective scheme for making use of implicit connectives. 5) At last, ``$\ell_2$-reg'' also implements feature mimicking by imposing an $\ell_2$ distance penalty between the implicit relation features and connective-augmented features. We see that the simple model has obtained improvement over previous best-performing systems in both settings, further validating the idea of imitation. However, in contrast to the fixed $\ell_2$ regularization, our adversarial framework provides an adaptive mechanism, which is more flexible and performs better as shown in the table.

\begin{table}[!h]
\small
	\centering
	\begin{tabular}{l  cccc}
	\cmidrule[\heavyrulewidth]{1-5}
	Model &\textsc{Comp.}&	\textsc{Cont.}&	\textsc{Exp.}& 	\textsc{Temp.}
\\ 
		\cmidrule{1-5}
        \newcite{pitler-louis-nenkova:2009:ACLIJCNLP}
		&21.96  
		&47.13
		&-
		&16.76\\
        \newcite{qin2016stacking}
        &{\bf 41.55}
        &{\bf 57.32}
        &71.50
        &35.43\\
        \newcite{D16-1037}
        &35.88
        &50.56
        &71.48
        &29.54\\
        \newcite{zhou2010predicting} 
        &31.79
		&47.16
		&70.11
		&20.30\\
        \newcite{D16-1130} 
		&36.70
		&54.48
		&70.43
		&{\bf 38.84}\\
	    \newcite{chen2016implicit} 
		&40.17
		&54.76
		&-
		&31.32\\
        \cmidrule{1-5}
		Ours
		&40.87
		&54.56
		&{\bf 72.38}
		&36.20\\      
		\cmidrule[\heavyrulewidth]{1-5}
	\end{tabular}
	\caption{Comparisons of $F_1$ scores (\%) for binary classification. Please see the supplementary materials for more details of the experiment setting.} 
    \label{tab:binary}
\end{table}

\begin{figure*}[!h]
	\centering
    \begin{subfigure}
		\centering
		\includegraphics[width=0.95\columnwidth]{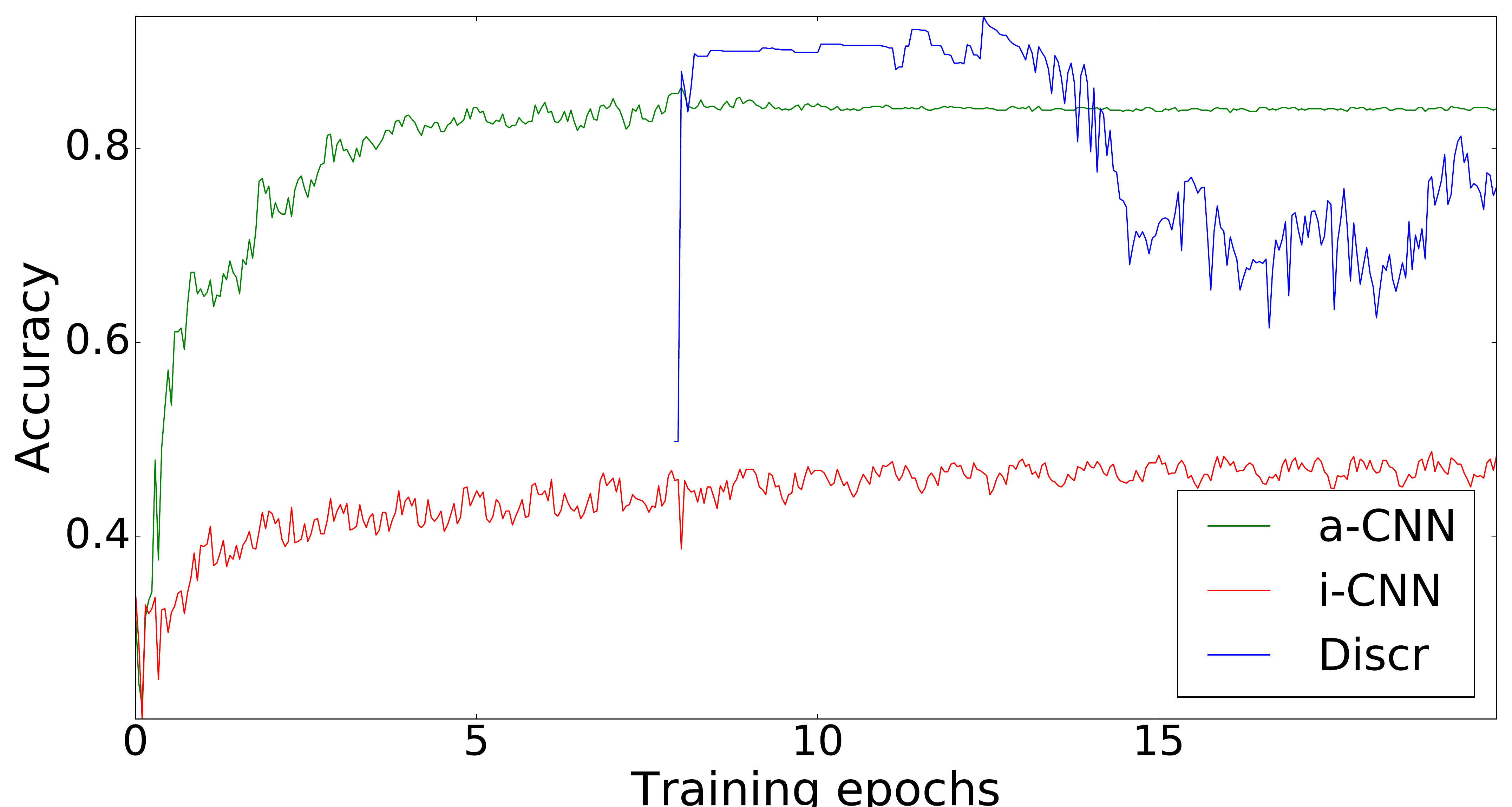}
	\end{subfigure}
    \begin{subfigure}
		\centering
		\includegraphics[width=0.95\columnwidth]{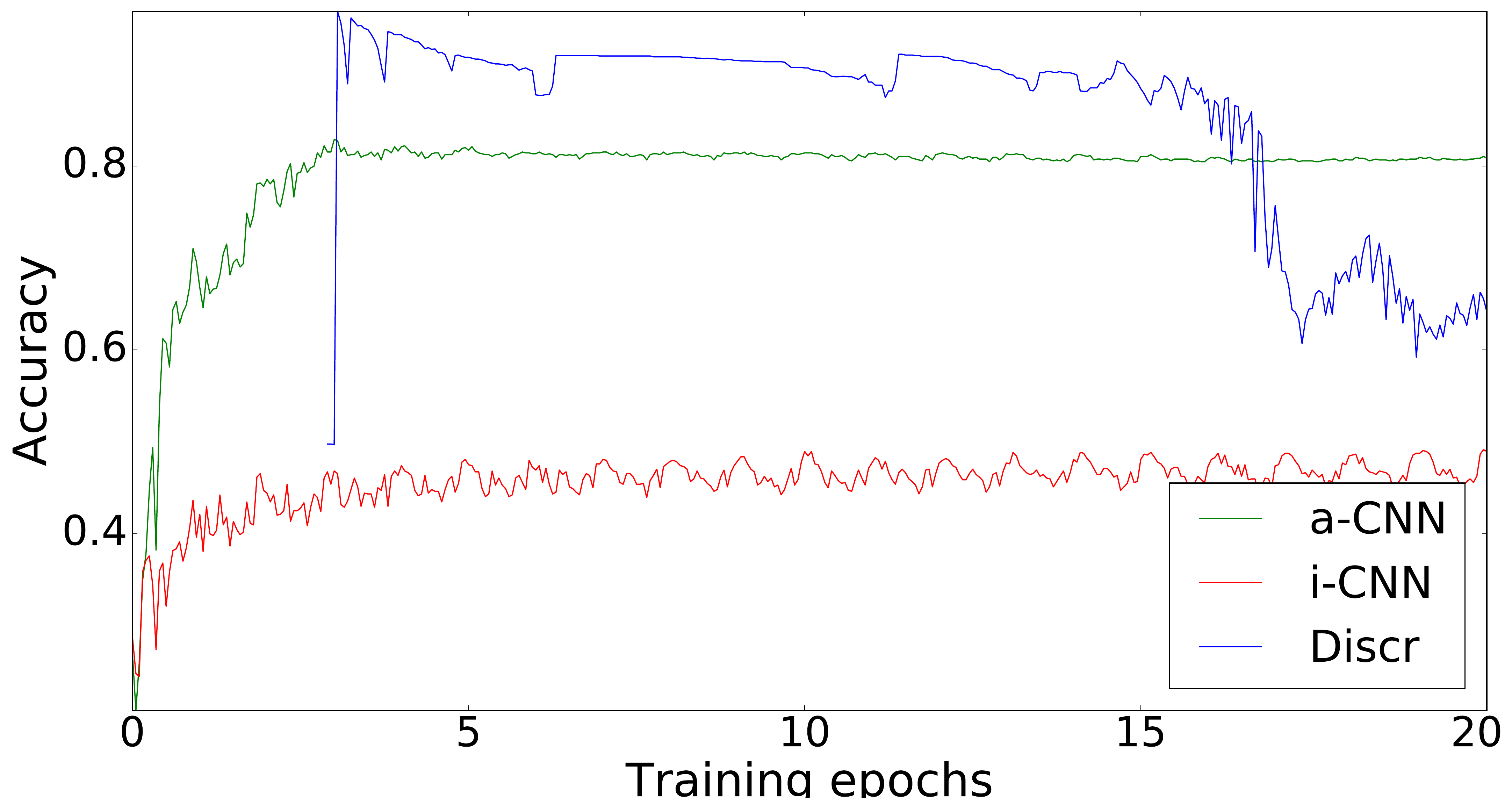}
	\end{subfigure}
	\caption{(Best viewed in colors.) Test-set performance of three components over training epochs. Relation networks {\it a-CNN} and {\it i-CNN} are measured with multi-class classification accuracy (with or without implicit connectives, respectively), while the discriminator is evaluated with binary classification accuracy. {\bf Top}: the PDTB-Lin setting~\cite{lin2009recognizing}, where first 8 epochs are for initialization stage (thus the discriminator is fixed and not shown); {\bf Bottom}: the PDTB-Ji setting~\cite{ji2015one}, where first 3 epochs are for initialization.}
	\label{fig:lines}
\end{figure*}
\begin{figure*}[!h]
	\centering
    \begin{subfigure}{(a)}
		\centering
		\includegraphics[width=0.5\columnwidth]{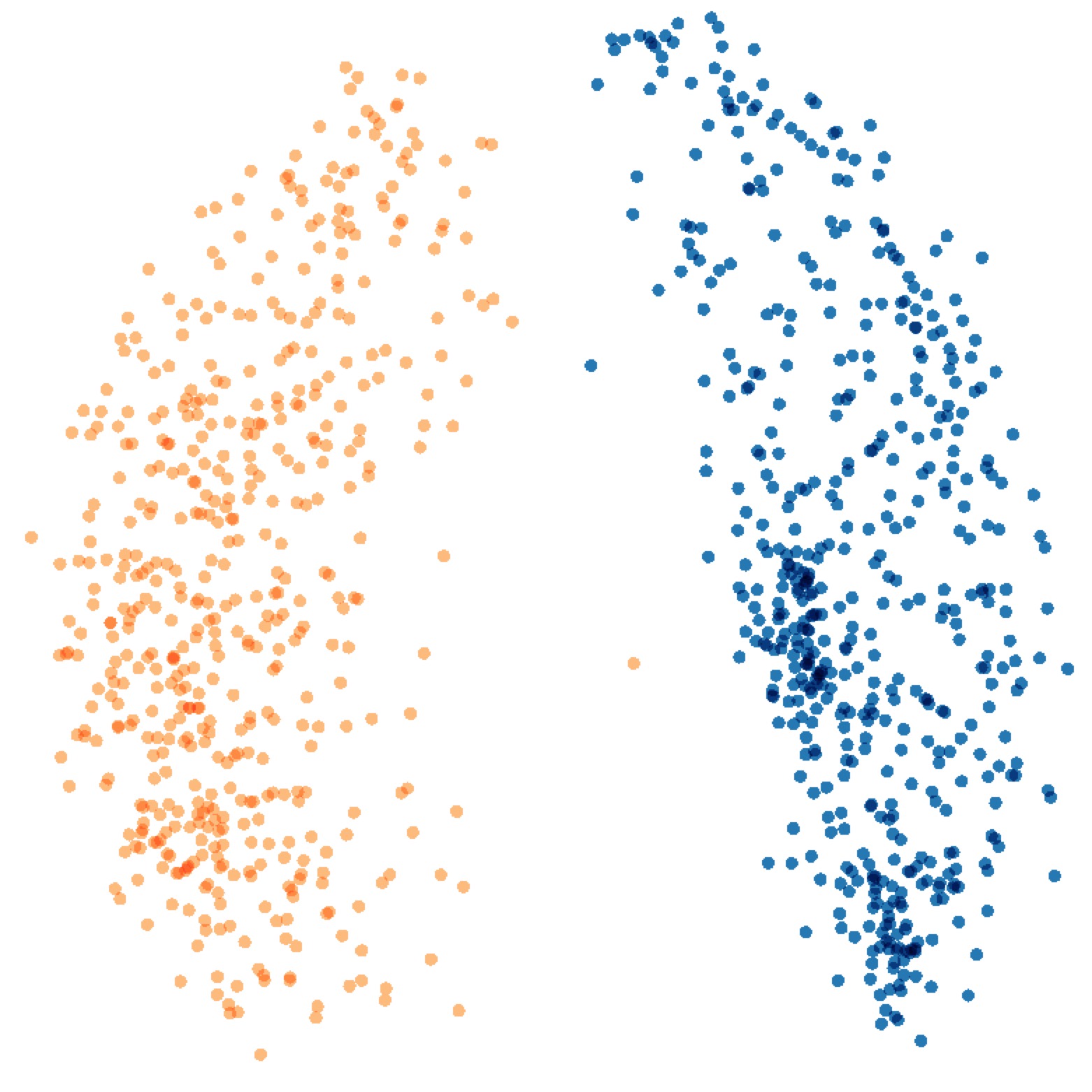}
	\end{subfigure}{(b)}
        \begin{subfigure}
		\centering
		\includegraphics[width=0.52\columnwidth]{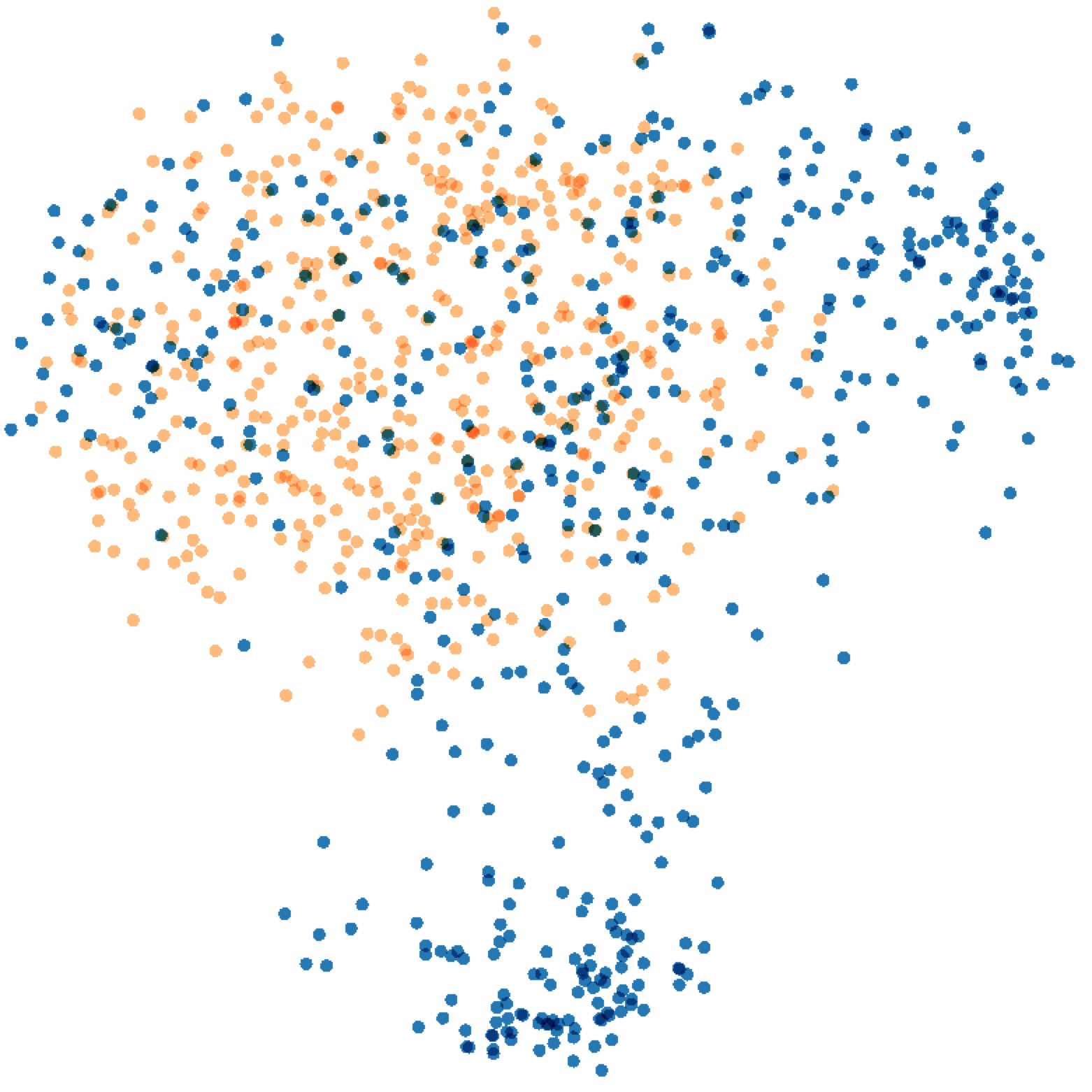}
	\end{subfigure}
    \begin{subfigure}{(c)}
		\centering
		\includegraphics[width=0.57\columnwidth]{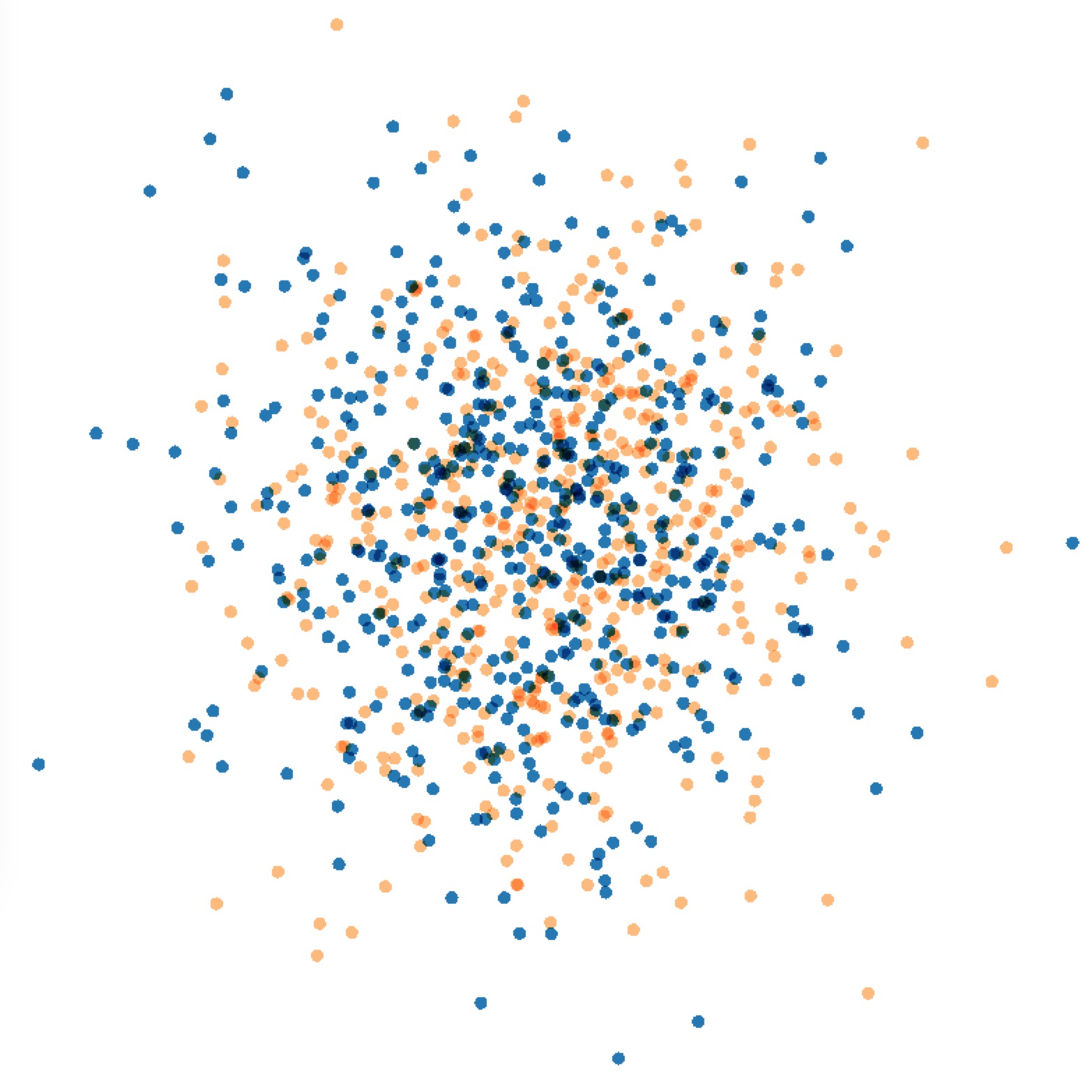}
	\end{subfigure}
	\caption{(Best viewed in colors.) Visualizations of the extracted hidden features by the implicit relation network {\it i-CNN} (blue) and connective-augmented relation network {\it a-CNN} (orange), in the multi-class classification setting~\cite{lin2009recognizing}. {\bf (a)} Two networks are trained without adversary (with shared classifier); {\bf (b)} Two networks are trained within our framework at epoch 10; {\bf (c)} Features at epoch 20. The implicit relation network successfully imitates the connective-augmented features through the adversarial game. Visualization is conducted with the t-SNE algorithm~\cite{maaten2008visualizing}.}
	\label{fig:viz}
\end{figure*}

\paragraph{One-versus-all Classifications}\quad\\
We also report the results of four one-versus-all binary classifications for more comparisons with prior work. We follow the conventional experimental setting~\cite{pitler-louis-nenkova:2009:ACLIJCNLP} by selecting sections 2-20, 21-22, and 0-1 as training, dev, and test sets. More detailed data statistics are provided in the supplementary materials.

Following previous work, the F1 scores are reported in Table~\ref{tab:binary}. Our method outperforms most of the prior systems in all the tasks. We achieve state-of-the-art performance in recognition of the Expansion relation, and obtain comparable scores with the best-performing methods in each of the other relations, respectively. Notably, our feature imitation scheme greatly improves over \cite{zhou2010predicting} which leverages implicit connectives as an intermediate prediction task. This provides additional evidence for the effectiveness of our approach.

\subsection{Qualitative Analysis}
We now take a closer look into the modeling behavior of our framework, by investigating the process of the adversarial game during training, as well as the feature imitation effects.

Figure~\ref{fig:lines} demonstrates the training progress of different components. The {\it a-CNN} network keeps high predictive accuracy as implicit connectives are given, showing the importance of connective cues.
The rise-and-fall patterns in the accuracy of the discriminator clearly show its competition with the implicit relation network {\it i-CNN} as training goes. At first few iterations the accuracy of the discriminator increases quickly to over $0.9$, while at late stage the accuracy drops to around $0.6$, showing that the discriminator is getting confused by {\it i-CNN} (an accuracy of $0.5$ indicates full confusion). 
The {\it i-CNN} network keeps improving in terms of implicit relation classification accuracy, as it is gradually fitting to the data and simultaneously learning increasingly discriminative features by mimicking {\it a-CNN}. The system exhibits similar learning patterns in the two different settings, showing the stability of the training strategy.

We finally visualize the output feature vectors of {\it i-CNN} and {\it a-CNN} using the t-SNE method~\cite{maaten2008visualizing} in Figure~\ref{fig:viz}. Without feature imitation, the extracted features by the two networks are clearly separated (Figure~\ref{fig:viz}(a)). In contrast, as shown in Figures~\ref{fig:viz}(b)-(c), the feature vectors are increasingly mixed as training proceeds. Thus our framework has successfully driven {\it i-CNN} to induce similar representations with {\it a-CNN}, even though connectives are not present.

\section{Discussions}\label{sec:conclude}
We have developed an adversarial neural framework that facilitates an implicit relation network to extract highly discriminative features by mimicking a connective-augmented network. Our method achieved state-of-the-art performance for implicit discourse relation classification. Besides implicit connective examples, our model can naturally exploit enormous explicit connective data to further improve discourse parsing.

The proposed adversarial feature imitation scheme is also generally applicable to other context to incorporate indicative side information available at training time for enhanced inference. Our framework shares a similar spirit of the iterative knowledge distillation method~\cite{hu2016harnessing,hu2016deep} which train a ``student'' network to mimic the classification behavior of a knowledge-informed ``teacher'' network. Our approach encourages imitation on the feature level instead of the final prediction level. This allows our approach to apply to regression tasks, and more interestingly, the context in which the student and teacher networks have different prediction outputs, e.g., performing different tasks, while transferring knowledge between each other can be beneficial. Besides, our adversarial mechanism provides an adaptive metric to measure and drive the imitation procedure.

\balance
\bibliography{acl2016}
\bibliographystyle{acl_natbib}

\clearpage

\onecolumn
\begin{appendices}

\numberwithin{equation}{section}

\section{Model architectures and training configurations}
In this section we provide the detailed architecture configurations of each component we used in the experiments.

\begin{itemize}
\item Table~\ref{tab:cnn-arch} lists the filter configurations of the convolutional layer in {\it i-CNN} and {\it a-CNN} in different tasks, tuned on dev sets. As described in section~3.3 in the paper, following the convolutional layer is a max pooling layer in {\it i-CNN}, and an average k-max pooling layer with $k=2$ in {\it a-CNN}. 
\item The final single-layer classifier {\it C} contains 512 neurons and uses ``tanh'' as activation function. 
\item The discriminator {\it D} consists of 4 fully-connected layers, with 2 gated pathways from layer 1 to layer 3 and layer 4 (see Figure~3 in the paper). The size of each layer is set to 1024 and is fixed in all the experiments.
\item We set the dimension of the input word vectors to 300 and initialize with pre-trained word2vec~\cite{mikolov2013distributed}. The maximum length of sentence argument is set to 80, and truncation or zero-padding is applied when necessary.
\end{itemize}

\begin{table*}[!h]
  \centering
  \begin{tabular}{|c|l|l|}
     \hline
     Task &  filter sizes & filter number   \\ \hline\hline
     {PDTB-Lin} & 2, 4, 8 & 3$\times$256 \\
	 \hline
     {PDTB-Ji} & 2, 5, 10 & 3$\times$256\\
	 \hline
     {One-vs-all} & 2, 5, 10 & 3$\times$1024\\
     \hline
   \end{tabular}
  \caption{The convolutional architectures of {\it i-CNN} and {\it a-CNN} in different tasks (section~4). For example, in PDTB-Lin, we use 3 sets of filters, each of which is of size 2, 4, and 8, respectively; and each set has 256 filters.}
  \label{tab:cnn-arch}
\end{table*}

All experiments were performed on a Linux machine with eight 4.0GHz CPU cores and 32GB RAM. We implemented neural networks based on Tensorflow\footnote{https://www.tensorflow.org}, a popular deep learning platform.

\section{One-vs-all Classifications}
Table~\ref{tab:binary-data} lists the statistics of the data.
\begin{table}[!h]
  \centering
  \normalsize
\begin{tabular}{l l l l } 
\hline
Relation &Train&Dev&Test\\ 
\hline
Comparison&1942/1942&197/986&152/894\\
Contigency&3342/3342&295/888&279/767\\
Expansion&7004/7004&671/512&574/472\\
Temporal&760/760&64/1119&85/961\\
\hline
\end{tabular}
	\caption{Distributions of positive and negative instances from the train/dev/test sets in four binary relation classification tasks.} 
\label{tab:binary-data}
\end{table}

\end{appendices}

\end{document}